%% file: bmvc_arxiv.tex
\documentclass{bmvc2k}

\usepackage{wrapfig}
\usepackage{amsmath,graphicx,caption}
\usepackage{amssymb}
\usepackage{xcolor}
\definecolor{bmvcbluecaption}{rgb}{0,0,.4} 
\title{Detecting Audio-Visual Deepfakes with Fine-Grained Inconsistencies}

\addauthor{Marcella Astrid}{marcella.astrid@uni.lu}{1}
\addauthor{Enjie Ghorbel}{enjie.ghorbel@isamm.uma.tn}{1,2}
\addauthor{Djamila Aouada}{djamila.aouada@uni.lu}{1}

\addinstitution{
 Computer Vision, Imaging \& Machine Intelligence Research Group (CVI$^2$),\\
 Interdisciplinary Centre for Security, Reliability and Trust (SnT),\\
 University of Luxembourg,\\
 Luxembourg 
}
\addinstitution{
 Cristal Laboratory,\\ 
 National School of Computer Sciences,\\
 Manouba University,\\
 Tunisia
}

\runninghead{Astrid \etal}{Detecting Audio-Visual Deepfakes with Fine-Grained ...}

\def\etal{\emph{et al}\bmvaOneDot}

\begin{document}

\maketitle

\begin{abstract}
Existing methods on audio-visual deepfake detection mainly focus on high-level features for modeling inconsistencies between audio and visual data. As a result, these approaches usually overlook finer audio-visual artifacts, which are inherent to deepfakes. Herein, we propose the introduction of fine-grained mechanisms for detecting subtle artifacts in both spatial and temporal domains. 
First, we introduce a local audio-visual model capable of capturing small spatial regions that are prone to inconsistencies with audio. For that purpose, a fine-grained mechanism based on a spatially-local distance coupled with an attention module is adopted.
Second,  we introduce a temporally-local pseudo-fake augmentation to include samples incorporating subtle temporal inconsistencies in our training set.  
Experiments on the DFDC and the FakeAVCeleb datasets demonstrate the superiority of the proposed method in terms of generalization as compared to the state-of-the-art under both in-dataset and cross-dataset settings.
\end{abstract}

\section{Introduction}
\label{sec:intro}
In recent years, the capabilities of generative techniques, especially deep learning-based methods, in creating audio-visual deepfake data have rapidly improved~\cite{jia2018transfer,korshunova2017fast,nirkin2019fsgan}. Despite their advantages, these techniques can also be damaging to society if used with malicious intent. For example, a finance worker was recently scammed out of 25 million US dollars after engaging in a video call with a deepfake of the company's chief financial officer~\cite{Scam_news}. Therefore, it is essential to propose detection methods that are capable of identifying deepfakes.




To detect audio-visual deepfakes, one can exploit the inconsistencies between audio and visual data present in deepfakes. Several methods have been developed by focusing on this aspect~\cite{chugh2020not, gu2021deepfake, feng2023self}. Nevertheless, despite their effectiveness, these methods have not explored fine-grained audio-visual representations. In fact, they only operate on high-level global features, while deepfake artifacts are known to be typically localized~\cite{nguyen2024laa,zhao2021multi}.
In this work, we posit that by employing fine-grained strategies for modeling subtle audio-visual artifacts, deepfakes can be detected more effectively.

\begin{figure}[tb]
\centering
\begin{minipage}{.50\textwidth}
  \centering
  \includegraphics[width=\linewidth]{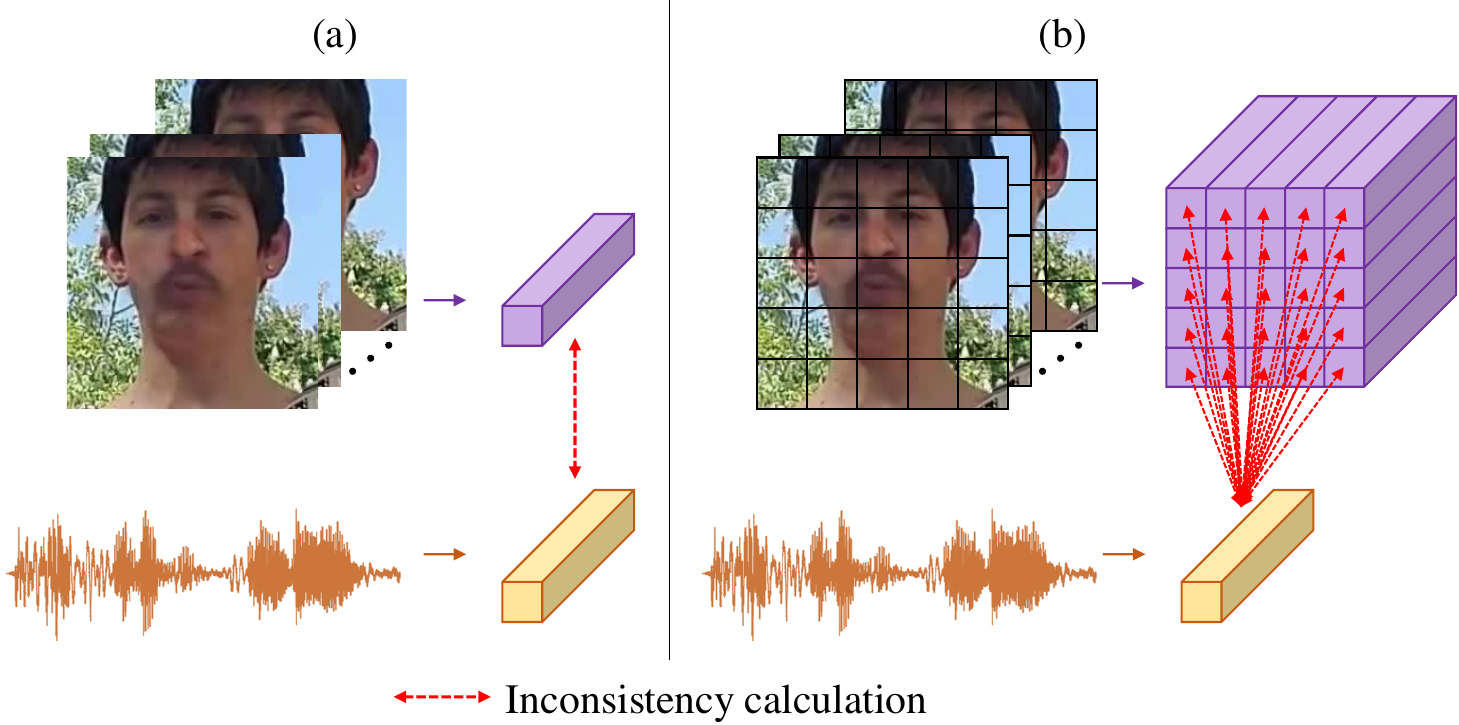}
  \vspace{3mm}
  \caption{(a) Previous works~\cite{chugh2020not,gu2021deepfake} utilize high-level global features to measure inconsistencies between audio and visual data. (b) The proposed method measures the inconsistency between  different visual regions and the audio input.}
  \label{fig:introduction}
\end{minipage}%
\hspace{0.04\textwidth}%
\begin{minipage}{.45\textwidth}
  \centering
  \includegraphics[width=\linewidth]{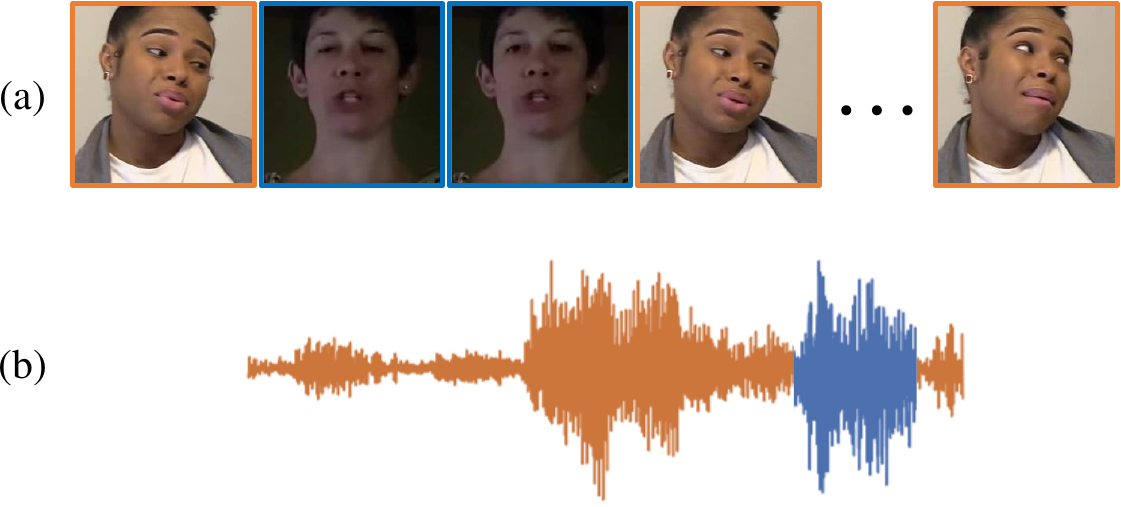}
  \vspace{3mm}
  \caption{(a) The proposed temporally-local pseudo-fake synthesis involves the replacement of a small video segment by a subsequence extracted from another video (marked in blue). (b) The same strategy is followed for audio data.}
  \label{fig:introduction_temporal}
\end{minipage}
\end{figure}

Based on this assumption, we introduce novel audio-visual fine-grained mechanisms tailored to deepfake detection at both the spatial and the temporal levels. 
To the best of our knowledge, no work has explored fine-grained inconsistencies to detect audio-visual deepfakes.
In the spatial domain, instead of measuring inconsistencies through high-level representations as done in previous works \cite{chugh2020not,gu2021deepfake} (see Figure \ref{fig:introduction}(a)), we consider features extracted from different spatial patches (Figure \ref{fig:introduction}(b)). Attention is also incorporated, as we believe that only specific regions of the spatial domain are relevant to deepfake detection. In the temporal domain, we propose augmenting the fake data by simulating audio-visual inconsistencies. However, instead of applying augmentations to the entire audio-visual stream as in~\cite{feng2023self}, we only manipulate a small portion of the sequence to simulate subtle artifacts, as illustrated in Figure~\ref{fig:introduction_temporal}. We conduct experiments on two audio-visual deepfake detection benchmarks, namely, DFDC and FakeAVCeleb. The results demonstrate enhanced generalization capabilities as compared to the state-of-the-art under both in-dataset and cross-dataset settings.

In summary, our contributions are listed below: 1) We propose a spatially-local architecture for detecting audio-visual deepfakes by implicitly measuring inconsistencies between sub-regions of the visual data and the audio; 
2) We apply an additional cross-attention mechanism that enforces the model to focus on the inconsistency-prone visual regions;
3) We propose a temporally-local pseudo-fake augmentation that simulates fine-grained audio-visual artifacts; 4) We train our model on the DFDC dataset~\cite{dolhansky2020deepfake} and evaluate it for both in-dataset (DFDC) and cross-dataset (FakeAVCeleb dataset~\cite{khalid2021fakeavceleb}) settings, demonstrating enhanced generalization capabilities in comparison to state-of-the-art (SoA) methods.

\noindent\textbf{Paper organization: } Section \ref{sec:relatedworks} discusses related work and positions our work with respect to the existing literature. In Section \ref{sec:methodology}, we present a detailed description of the proposed method. Section \ref{sec:experiments} outlines the experimental setup and presents the obtained results. Finally, Section \ref{sec:conclusion} summarizes our findings and concludes this paper.

\section{Related work}
\label{sec:relatedworks}

\subsection{Audio-visual deepfake detection}

There exist numerous approaches for detecting audio-visual deepfakes. One approach focuses on specific identities, targeting the detection of deepfakes related to certain individuals~\cite{agarwal2023watch,cozzolino2023audio,cheng2022voice}. While useful for some applications, these methods are limited to detecting deepfakes of individuals included in the training set. In contrast, we aim for a more universal approach capable of detecting individual-agnostic deepfakes. Another approach integrates information from both audio and visual modalities~\cite{yang2023avoid,wang2024avt2,zhou2021joint,salvi2023robust,lewis2020deepfake,korshunov2018speaker,korshunov2019tampered,lomnitz2020multimodal}. However, combining features may induce redundancy, which has been shown to compromise the generalization capabilities~\cite{nguyen2024laa}. This is also confirmed by our experiments where the use of residual connections to inject redundant information has led to a decrease in performance. A third approach for multi-modal deepfake detection is to aggregate the predictions resulting from separate visual and audio models~\cite{ilyas2023avfakenet}. Then, if at least one model predicts the input as fake, the overall prediction is set to fake. However, this mechanism may lead to a large number of false positives. One method in deepfake detection utilizes audio and visual modalities to pretrain a model~\cite{haliassos2022leveraging}. However, the final model is fine-tuned only on visual data and hence cannot be used for audio-visual deepfake detection.

Different from the aforementioned methods, our work is primarily related to approaches that exploit inconsistencies between audio and visual cues present in fake data. Existing methods have addressed this issue in numerous ways. Chugh \etal \cite{chugh2020not} aim to minimize the distance between audio and visual features extracted from real data while maximizing it otherwise. However, this matching primarily operates on high-level features, potentially overlooking subtle artifacts. Similarly, Gu \etal \cite{gu2021deepfake} restrict their analysis to the lips only, neglecting potential artifacts elsewhere on the face. Mittal \etal \cite{mittal2020emotions} focus on detecting mismatches in high-level emotions between visual and audio. Nevertheless, this approach heavily relies on emotion recognition models, which could be sub-optimal for deepfake detection. Feng \etal \cite{feng2023self} employ an auto-regressive model to predict the synchronization of visual and audio pairs. Nonetheless, their approach involves the translation of the entire audio/visual input to mimic inconsistencies, potentially disregarding more localized discrepancies. In contrast, we propose a fine-grained method for detecting subtle audio-visual mismatches taking into account both the spatial and the temporal dimensions.

\subsection{Pseudo-fake generation}

Pseudo-fake generation has been widely acknowledged for its ability to enhance the generalization capabilities of deepfake detectors. In the visual domain, Li \etal \cite{li2020face} blend a face into another facial image. To reduce blending artifacts, Shiohara and Yamasaki \cite{shiohara2022detecting} blend faces from the same individual, demonstrating significant improvements in generalization as compared to \cite{li2020face}. Mejri \etal \cite{mejri2023untag} focus on blending specific facial regions, such as eyes or nose. Chen \etal \cite{chen2022self} employ reinforcement learning to generate pseudo-fakes. In the temporal domain, Wang \etal \cite{wang2023altfreezing} introduce temporal inconsistencies by dropping or repeating frames. In the audio-visual domain, Feng \etal \cite{feng2023self} translate entire audios or image sequences to simulate inconsistencies. In this work, inspired by the findings of \cite{shiohara2022detecting}, we explore the generation of subtle pseudo-fake synthesis, but consider the simulation of audio-visual inconsistencies in the temporal domain.

\subsection{Fine-grained deepfake detection}

While simple binary deep neural networks have been initially employed for deepfake detection \cite{rossler2019faceforensics++}, they usually struggle to capture localized artifacts \cite{zhao2021multi}. To overcome this limitation,
several strategies have been proposed for visual deepfake detection.
For instance, Zhao \etal \cite{zhao2021multi} utilize multiple attention modules to capture fine-grained features across different regions of the input. On the other hand, Nguyen \etal \cite{nguyen2024laa} guide explicitly the network to focus on vulnerable points that are defined as the pixels which potentially suffer the most from blending artifacts. Nevertheless, to the best of our knowledge, no prior investigation has focused on examining inconsistencies between audio and local regions of the visual data for detecting audio-visual deepfakes. 

\section{Methodology}
\label{sec:methodology}

\begin{figure*}[]
\centering
\includegraphics[width=\linewidth]{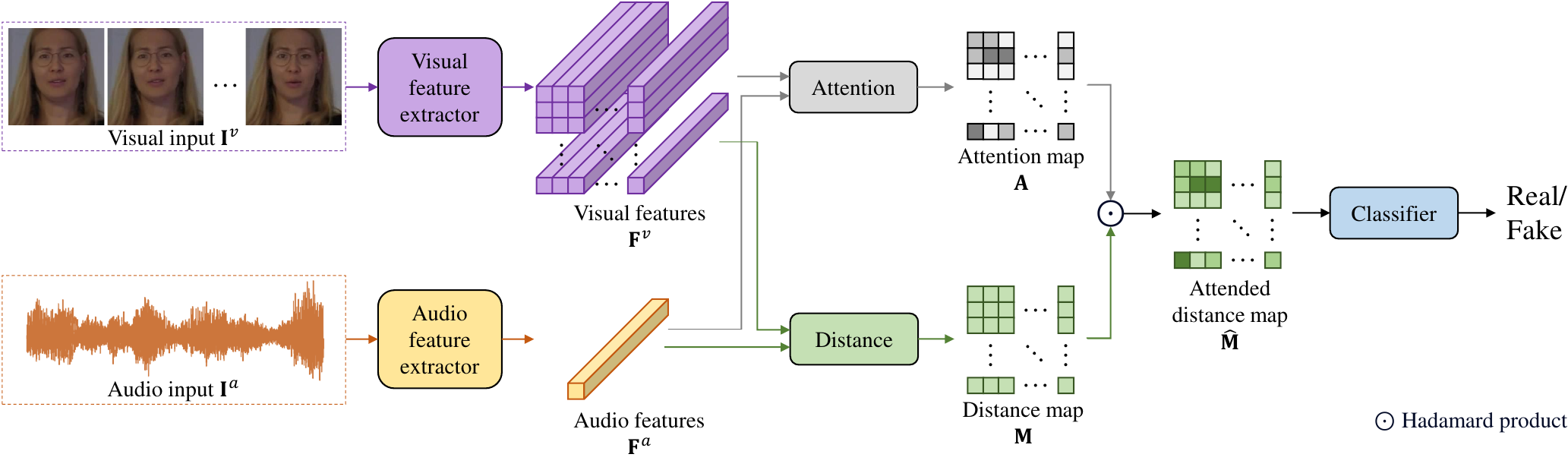}
\vspace{1mm}
\caption{The proposed spatially-local deepfake detector: Firstly, audio and visual features are extracted, separately. Next, we compute the distance and attention maps between the audio and all spatial positions of the visual features. Subsequently, the distance map and the attention map are multiplied before being fed into a single-layer real/fake classifier.}
\label{fig:method}
\end{figure*}

In order to focus on subtle artifacts, we propose two components: a spatially-local  architecture for audio-visual deepfake detection (Section \ref{subsec:spatially-local}) and a temporally-local pseudo-fake data generation (Section \ref{subsec:temporally-local}).

\subsection{Spatially-local audio-visual deepfake detector architecture}
\label{subsec:spatially-local}
Figure \ref{fig:method} illustrates the overall architecture of the proposed deepfake detector, which extracts the spatially-local inconsistencies map to classify whether an input pair is fake or not. It comprises the feature extractor, the distance calculation, the attention module, and the classifier.

\subsubsection{Feature extractor}

The input to our feature extractor is a pair of audio and visual inputs, denoted as  $\mathbf I^a$ and $\mathbf I^v$, respectively. The audio is represented as a waveform of size $T^{^a} \times C^{a} = 48000 \times 1$, where $T^{a}$ and $C^{a}$ denote the temporal and channel dimensions, respectively. We opt for the waveform representation to mitigate dependencies on frequency conversion processes, which can potentially reduce the robustness of the detector \cite{tak2021end,jung2019rawnet}. The visual input consists of a sequence of images of size $T^{v} \times C^{v} \times H^{v} \times W^{v} = 30 \times 3 \times 224 \times 224$, where $T^{v}$, $C^{v}$, $H^{v}$, and $W^{v}$ represent the temporal (number of frames), the channel, the height, and the width dimensions, respectively.

Each input is fed into a specialized feature extractor denoted as $\mathcal{A}(\cdot)$ and $\mathcal{V}(\cdot)$ as follows,
\begin{equation}
   \mathbf F^a = \mathcal{A}(\mathbf I^a) \text{, } \quad \mathbf F^v = \mathcal{V}(\mathbf I^v) \text{,}  
\end{equation}
where $\mathbf F^a$ represents the audio feature of size $T^{a \prime}\times C^{a\prime} = 128 \times 15$ and $\mathbf F^v$ represents the visual feature of size $T^{v\prime} \times C^{v\prime} \times H^{v\prime} \times W^{v\prime} = 128 \times 15 \times 28 \times 28$. 

The feature extractor $\mathcal{V}(\cdot)$ is based on ResNetConv3D~\cite{hara2018can} that we modify to be shallow in order to obtain high-resolution features. Moreover, recent works in visual deepfake detection also suggest that visual deepfake artifacts can be effectively captured with a shallower network \cite{mejri2021leveraging,afchar2018mesonet}. The branch $\mathcal{A}(\cdot)$ is then based on a Conv1D architecture. It is designed such that $T^{a \prime}$ and $C^{a \prime}$ are equal to $T^{v \prime}$ and $C^{v \prime}$, respectively, so that we can calculate the distance between audio and visual features in the next step.





\subsubsection{Spatially-local distance map}
To measure fine-grained inconsistencies between the audio and different spatial visual patches, we create a distance map $\mathbf M=(M_{i,j})_{ 1\leq i\leq H^{v \prime} ,1\leq j\leq  W^{v \prime} }$, where each element $M_{i,j}$ is calculated as follows,
\begin{equation}
    M_{i,j} = d(\mathbf f^a, \mathbf f^v_{i,j}) \text{,}
\end{equation} 
where $d(\cdot)$ represents a distance function. In our experiments, $d(\cdot)$ corresponds to the $L2$ distance. $\mathbf f^a$ and $\mathbf f^v_{i,j}$ denote the flattened versions of $\mathbf F^a$ and $\mathbf F^v_{i,j}$ (i.e., $\mathbf F^v$ at position $(i, j)$), respectively. Both vectors $\mathbf f^a$ and $\mathbf f^v_{i,j}$ have a size of $T^{v\prime}.C^{v\prime}$. Currently, we use the $L2$ distance following as in \cite{chugh2020not} to measure the distance between global features of audio and visual data. Nevertheless, other distance functions can also be explored in the future.

\subsubsection{Attention module}
Some visual regions of $\mathbf M$, such as hair and background, might be irrelevant to the quantification of audio-visual inconsistencies. Therefore, we propose to utilize an attention map $\mathbf A$ of the same size as $\mathbf M$ to implicitly reduce the contributions of these regions in the final classification. This is achieved through the following element-wise multiplication,
\begin{equation}
    \hat{\mathbf M} = \mathbf M \odot \mathbf A \text{,}
\end{equation}
where $\odot$ represents the Hadamard/element-wise product.

The attention map $ \mathbf A=(A_{i,j})_{ 1\leq i\leq H^{v \prime} ,1\leq j\leq  W^{v \prime} }$ is calculated using a cross-attention-like mechanism as described below,
\begin{equation}
    A_{i,j} = s \left( \frac{\mathcal{E}^a(\mathbf F^a) \cdot (\mathcal{E}^v(\mathbf F^v))_{i,j}} {T^{v\prime}C^{v\prime}} \right) \text{,}
\end{equation}
where $s$ represents the softmax activation function, $\mathcal{E}^a$ is a Conv1D operation with $C^{v\prime}/4$ filters, $\mathcal{E}^v$ is a Conv3D operation also with $C^{v\prime}/4$ filters, and $\cdot$ denotes the dot product. Division by $T^{v\prime}C^{v\prime}$ is a normalization process. 


\subsubsection{Classifier}
Finally, the estimated distance map $\hat{\mathbf M}$ serves as the input to the classifier $\mathcal{C}$ as depicted in the following equation, 
\begin{equation}
    y = \mathcal{C}(\hat{\mathbf M}) \text{,}   
\end{equation}
where $y$ tends towards 1 if the input is classified as fake, indicating a likely presence of inconsistencies in $M$, and approaches 0 otherwise. The classifier $\mathcal{C}$ consists of flattening $\hat{\mathbf M}$, followed by a linear layer with a sigmoid activation function. Thus, $\mathcal{C}$ only use the information of local inconsistencies to predict whether an audio-visual input is fake. Training is performed using a binary cross-entropy loss.

\subsection{Temporally-local pseudo-fake synthesis}
\label{subsec:temporally-local}

In previous work \cite{feng2023self}, the entire sequence of either audio or visual data is replaced with an alternative content for generating pseudo-fakes. However, such a technique usually produces low-quality pseudo-fakes, thereby hindering the generalization capabilities of the network. To address this, we propose a data synthesis method that replaces only a local portion of the audio, the visual, or both, to create multi-modal pseudo-fakes incorporating more subtle inconsistencies. This approach is inspired by image deepfake detection techniques, where it has been demonstrated that the use of imperceptible pseudo-fakes enhances the generalization capabilities \cite{shiohara2022detecting} of deepfake detectors. Note that our method remains inherently different, as it focuses on subtle inconsistencies between audio and visual data; rather than visual artifacts.

Given two visual or audio sequences of length $n$ denoted as $\mathbf I = \{\mathbf I_1, \mathbf I_2, \mathbf  I_3, ..., \mathbf I_n\}$ and $\mathbf J = \{\mathbf J_1, \mathbf J_2, \mathbf J_3, ..., \mathbf J_n\}$,  we randomly select a segment of length $l$, which is randomly chosen within the range $[l_{min}, l_{max}]$, where $l_{min} \leq l_{max}$. The values of $l_{min}$ and $l_{max}$ are determined as follows,
\begin{equation}
    l_{min} = r_{min} . n \text{, } \qquad l_{max} = r_{max} . n \text{, }
\end{equation}
where $r_{min}$ and $r_{max}$ are hyperparameters representing the length ratios, selected from the range $]0, 1]$. In addition to providing the locality extent, such a range also offers data diversity.

\begin{figure}[]
\centering
\includegraphics[width=\linewidth]{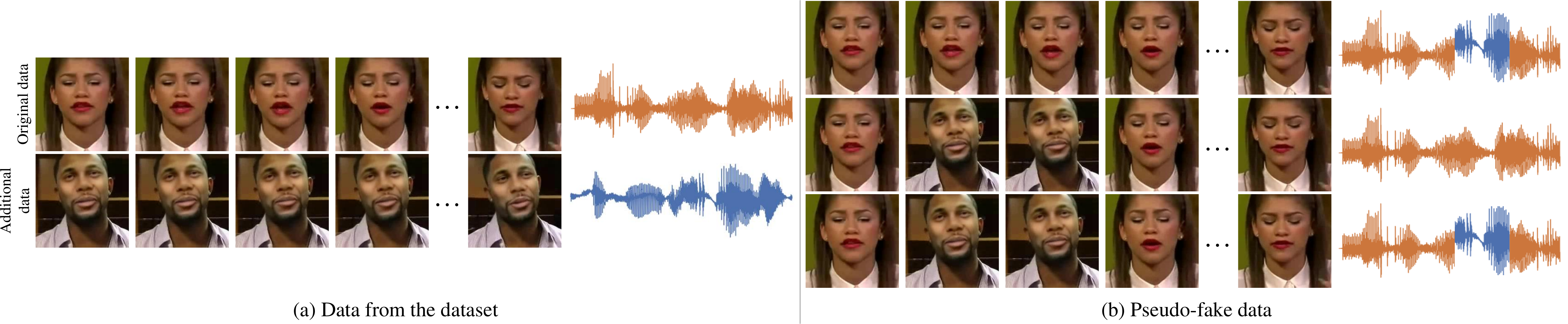}
\vspace{1mm}
\caption{Our temporally-local pseudo-fake data synthesis: Given the original dataset illustrated in (a), we can create three types of pseudo-fakes: modifying only the audio data, modifying only the visual data, or modifying both the audio and visual inputs, as illustrated in (b).}
\label{fig:local_pseudo_fake}
\end{figure}

After fixing  $l$, we randomly select a starting point  $g \in [1, n-l]$. Subsequently, we replace the elements from $\mathbf I_g$ to $\mathbf I_{g+l}$ with the corresponding elements $\mathbf J_g$ to $\mathbf J_{g+l}$. This replacement process can be applied to either visual data only, audio data only, or both. 
We incorporate the original data during training and dynamically synthesize pseudo-fakes with a probability of $0.5$. The pseudo-fake data is considered as fake during the learning phase. 
An illustration of the local replacement process is provided in Figure \ref{fig:local_pseudo_fake}.

\section{Experiments}
\label{sec:experiments}

\subsection{Experiment setup}


\noindent\textbf{Dataset.} We evaluate our method using the DFDC \cite{dolhansky2020deepfake} and the FakeAVCeleb \cite{khalid2021fakeavceleb} datasets. For DFDC, following the setup in \cite{chugh2020not,mittal2020emotions}, we sample $15,300$ training videos and $2,700$ test videos. In the training set, we ensure a balanced distribution between real and fake videos by randomly selecting the same number of videos for each class, while in the test set, we maintain a distribution that is identical to the original dataset. The testing protocol where we use the test split of DFDC is referred to as the \textit{in-dataset} setup. For the \textit{cross-dataset} setup, we utilize FakeAVCeleb solely during testing. We adopt the testing settings described in \cite{khalid2021evaluation}, where the test set comprises $70$ real and $70$ fake videos. We preprocess our dataset similarly to \cite{chugh2020not}, except that we use the audio waveform format normalized to a range of $-1$ to $1$ using min-max normalization. We train our model with DFDC, similar to the setup in \cite{chugh2020not, mittal2020emotions}. Nevertheless, training on FakeAVCeleb can also be explored in the future.

\noindent\textbf{Feature extractor architecture.} For the visual feature extractor $\mathcal{V}(\cdot)$, we modify the visual stream used in \cite{chugh2020not} by only using the first three convolution blocks. We build the audio feature extractor $\mathcal{A}(\cdot)$ using several Conv1D, batch norm, max pool, and ReLU layers.

\noindent\textbf{Evaluation criteria.} We assess the performance of our model using the video-level Area Under the ROC Curve (AUC). Video-level prediction is obtained by averaging the predictions of multiple input subsequences.


\noindent\textbf{Parameters and implementation details.} The model is trained using the Adam optimizer \cite{kingma2014adam} for 100 epochs, with a learning rate of $10^{-3}$, weight decay of $10^{-5}$, and a batch size of $8$. The model with the lowest loss across the 100 epochs is selected for testing. To ensure a wide variety of pseudo-fake samples, unless specified otherwise, we set $r_{min}$ to a value close to zero $\sim0$ (resulting in $l_{min} = 2$) and $r_{max}$ to 1.

\begin{figure}
  \begin{minipage}{.40\linewidth}
    \centering
    \resizebox{\linewidth}{!}{
    \begin{tabular}{|c|ccc|cc|}
      \hline
      \multicolumn{1}{|l|}{} & Att. & PF & RC & DFDC  & FAV   \\ \hline
      (a)                    &      &    &    & 97.11\% & 56.83\% \\
      (b)                    &      & \checkmark  &    & 94.47\% & \underline{71.24\%} \\
      (c)                    & \checkmark    &    &    & \textbf{98.09\%} & 58.57\% \\
      (d)                    & \checkmark    & \checkmark  &    & \underline{97.81\%} & \textbf{82.51\%} \\
      (e)                    & \checkmark    & \checkmark  & \checkmark  & 93.45\% & 60.27\% \\ \hline
    \end{tabular}
    }
    \vspace{2mm}
    \captionof{table}{Ablation study of our work reported in terms of AUC on DFDC and FakeAVCeleb (FAV) datasets. "Att.", "PF", and "RC" represent Attention, Pseudo-Fakes, and Residual Connections, respectively. The results produced by \textbf{our method} are reported in \textbf{(d)}. The best and second-best performances are marked with bold and underlined, respectively.}
    \label{tab:ablation}
  \end{minipage}\hfill
  \begin{minipage}{.55\linewidth}
    \centering
    \includegraphics[width=\linewidth]{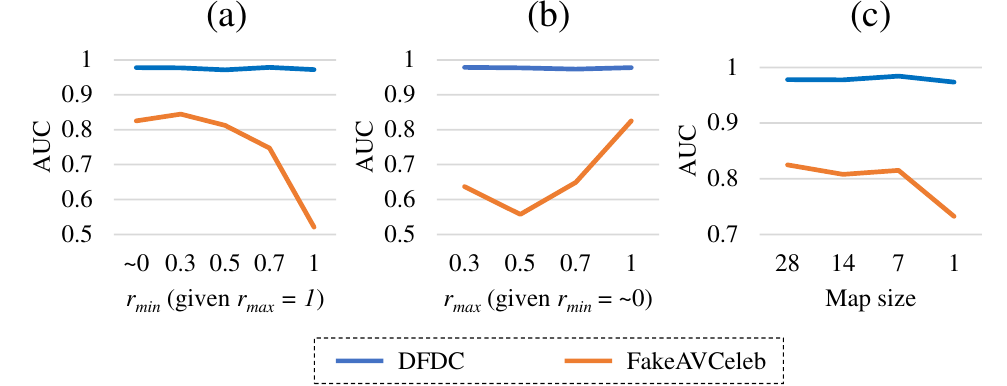}
    \vspace{2mm}
    \caption{AUC values under different (a) $r_{min}$, (b) $r_{max}$, and (c) map size ($H^{v\prime}$ and $W^{v\prime}$) settings on the DFDC dataset (in-dataset) and the FakeAVCeleb dataset (cross-dataset). Here, $r_{min} = \sim 0$ corresponds to $l_{min} = 2$.  In the in-dataset setting, no significant difference is observed, while the opposite is observed in the cross-dataset setting.}
    \label{fig:ablation_temporal_spatial}
  \end{minipage}
\end{figure}

\subsection{Ablation study}

\subsubsection{Architecture design}
For demonstrating the relevance of the proposed architecture, we conduct an ablation study and report the obtained results on DFDC and FakeAvCeleb with and without attention (i.e., $\hat{\mathbf M} = \mathbf M$) and  using a residual connection (i.e., $\hat{\mathbf M} = (\mathbf M \odot \mathbf A) + \mathbf M$) to simulate the effect of  introducing redundant information to the classifier. 

\begin{figure}[tb]
\centering
\begin{minipage}{.47\textwidth}
  \centering
  \includegraphics[width=\linewidth]{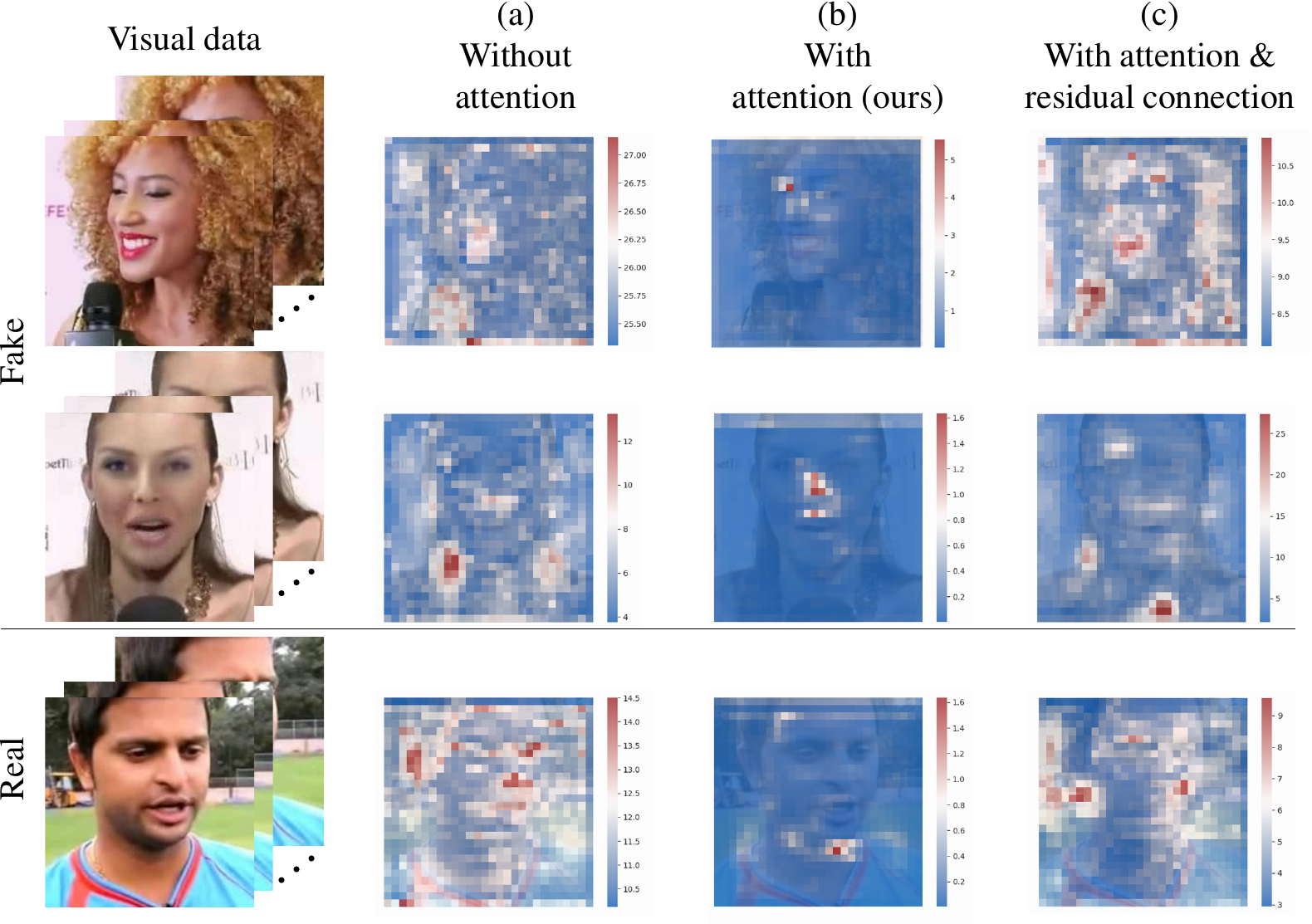}
  \vspace{2mm}
  \captionof{figure}{Visualization of $\hat{\mathbf M}$ extracted from three samples of the FakeAVCeleb dataset. Figures (a), (b), and (c) correspond to the same settings used in Table \ref{tab:ablation}(b), (d), and (e), respectively. When the proposed attention mechanism is used without residual connections, the map is more focused on specific parts as compared to other settings.}
  \label{fig:ablation_qualitative}
\end{minipage}%
\hfill
\begin{minipage}{.50\textwidth}
  \centering
  \includegraphics[width=0.75\linewidth]{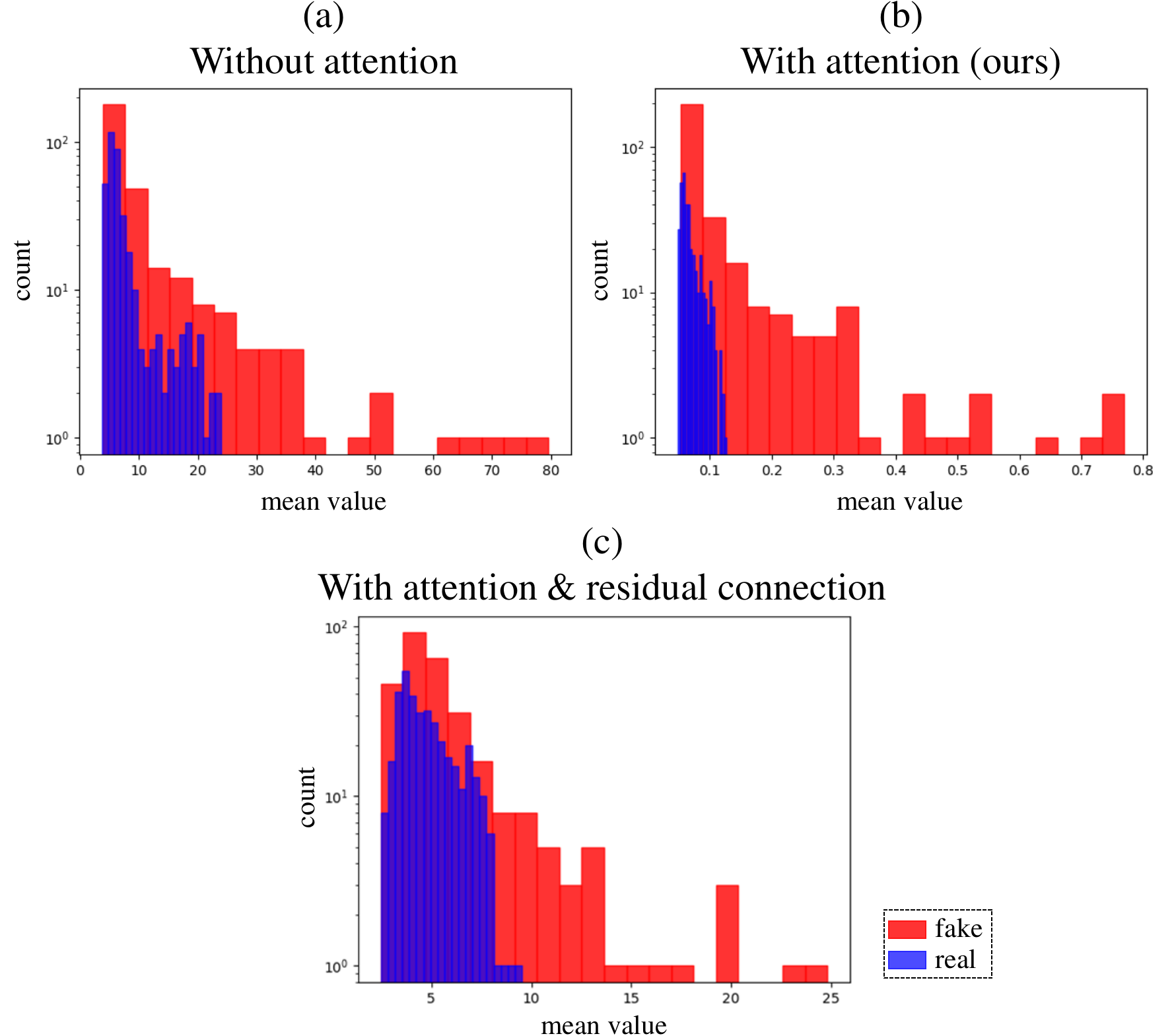}
  \vspace{2mm}
  \captionof{figure}{Histograms illustrating the distribution of the mean value of $\hat{\mathbf M}$ for real and fake data from the FakeAVCeleb test split. In Figures (a), (b), and (c), the same settings used in Table \ref{tab:ablation}(b), (d), and (e) are considered, respectively. The separation between the distributions of real and fake data is more pronounced in Figure (b) as compared to Figures (a) and (c).}
  \label{fig:ablation_meandistance_distribution}
\end{minipage}
\end{figure}

Specifically, Table \ref{tab:ablation}(b), (d), and (e) present the results of the model without attention, with attention, and with attention and a residual connection, respectively. It can be observed that the best results are obtained in Table \ref{tab:ablation}(d)  under both in-dataset and cross-dataset settings. For a deeper analysis of the results, we visualize the map $\hat{\mathbf M}$ in Figure \ref{fig:ablation_qualitative}. In Figure \ref{fig:ablation_qualitative}(b), we can observe that by only using attention, the model is able to better focus on specific regions as compared to the other setups. This suggests that attention allows disregarding some irrelevant parts such as the background; therefore leading to a more effective model. This gain in performance is also confirmed in Figure \ref{fig:ablation_meandistance_distribution}, where the distance between audio and visual features extracted from a real video tend to be lower when leveraging attention only, resulting in better discrimination between fake and real samples.

Moreover, for analyzing the impact of the proposed map distance, we also conducted experiments with different map sizes ($H^{v\prime}$ and $W^{v\prime}$, where $H^{v\prime} = W^{v\prime}$). To achieve that, we have applied an adaptive average pooling to $\mathbf F^{v}$ before calculating the distance with the audio features, as applying global average pooling before classifier is a common practice in computer vision \cite{he2016deep}. Figure \ref{fig:ablation_temporal_spatial}(c) shows that even though there is generally no significant difference in performance when varying map sizes in the in-dataset setup, there is a significant drop in performance when the features become too global ($H^{v \prime} = W^{v \prime} = 1$) in the cross-dataset setting.

\begin{table}[tb]
\parbox[t]{.37\linewidth}{
\centering
\resizebox{\linewidth}{!}{
\begin{tabular}{|l|c|c|}
\hline
Method       & Modality & FakeAVCeleb \\ \hline
LipForensics \cite{haliassos2021lips} & V        & 49.2\%        \\
CViT \cite{wodajo2021deepfake}         & V        & 45.5\%       \\
MesoNet \cite{afchar2018mesonet}      & V        & 54.1\%        \\ \hline
MDS \cite{chugh2020not}          & AV       & 72.9\%        \\ 
AVoiD-DF \cite{yang2023avoid}     & AV       & 82.8\%        \\
AVT$^2$-DWF \cite{wang2024avt2}     & AV       & 77.2\%        \\
Ours         & AV       & \textbf{84.5\%}        \\ \hline
\end{tabular}
}
\vspace{2mm}
\caption{AUC under  the cross-dataset setting, i.e., training on DFDC and testing on FakeAVCeleb. "V" and "AV" refer to visual-only and audio-visual modalities, respectively. The best performance is marked with bold.}
\label{tab:cross_soa}
}
\hfill
\parbox[t]{.56\linewidth}{
\centering
\resizebox{\linewidth}{!}{
\begin{tabular}{|l|c|c||l|c|c|}
\hline
Method          & Modality & DFDC & Method    & Modality & DFDC \\ \hline
Multi-attention \cite{zhao2021multi} & V        & 84.8\% & MDS \cite{chugh2020not}       & AV       & 90.7\% \\
SLADD \cite{chen2022self}          & V        & 75.2\% & Emotion \cite{mittal2020emotions}  & AV       & 84.4\% \\
LipForensics \cite{haliassos2021lips}    & V        & 73.5\% & BA-TFD \cite{cai2022you}   & AV       & 84.6\% \\
CViT \cite{wodajo2021deepfake}           & V        & 63.7\% & AVFakeNet \cite{ilyas2023avfakenet} & AV       & 86.2\% \\
MesoNet \cite{afchar2018mesonet}         & V        & 75.3\% & VFD \cite{cheng2023voice}      & AV       & 85.1\% \\ \cline{1-3}
AASIST \cite{jung2022aasist}         & A        & 68.4\% & AvoiD-DF \cite{yang2023avoid}  & AV       & 94.8\% \\
ECAPA-TDNN \cite{desplanques2020ecapa}     & A        & 69.8\% & AVT$^2$-DWF \cite{wang2024avt2}  & AV       & 89.2\% \\
RawNet \cite{jung2019rawnet}         & A        & 56.2\% & Ours      & AV       & \textbf{97.7\%} \\ \hline
\end{tabular}
}
\vspace{2mm}
\caption{AUC under the in-dataset setting, i.e., training on DFDC and testing also on DFDC. "V", "A", and "AV"  refer to visual-only, audio-only, and audio-visual modalities, respectively. The best performance is marked with bold.}
\label{tab:in_soa}
}
\end{table}


\subsubsection{Pseudo-fakes}
In this subsection, we explore the relevance of the temporally-local pseudo-fake  synthesis as well as the influence of $r_{min}$ and $r_{max}$.

Specifically, we compare the results of the models without pseudo-fakes (Table \ref{tab:ablation}(a) and (c)) and to the ones trained on pseudo-fakes (Table \ref{tab:ablation}(b) and (d)). The substantial improvement observed in FakeAVCeleb suggests enhanced generalization capabilities when utilizing pseudo-fake data. 
While the performance on DFDC slightly decreases when incorporating pseudo-fakes, the drop in performance remains negligible as compared to the obtained improvement under the most relevant scenario i.e., the cross-dataset setting, especially when using attention module.

\begin{figure*}[]
\centering
\includegraphics[width=\linewidth]{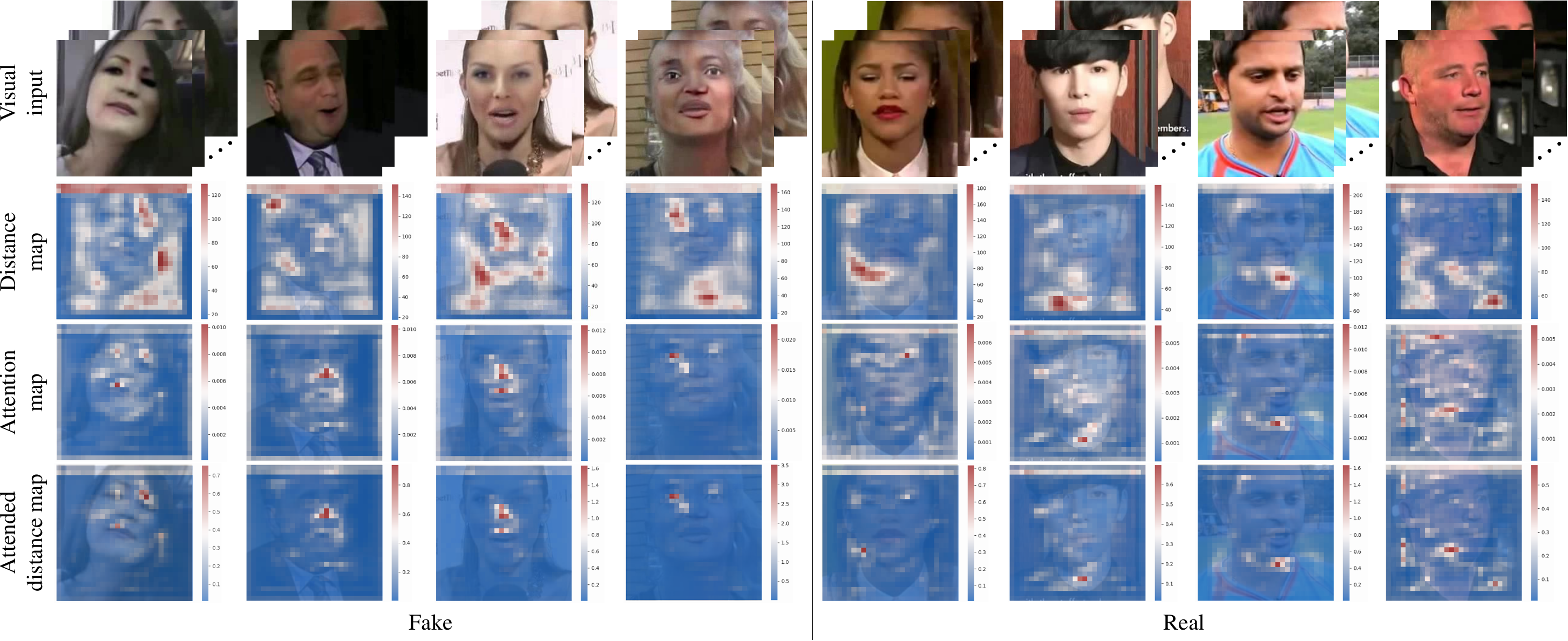}
\vspace{1mm}
\caption{Visualization of the distance map $\mathbf M$, the attention map $\mathbf A$, and the attention-aware distance map $\hat{\mathbf M}$ for several examples from the FakeAVCeleb dataset. It is observed that $\mathbf A$ reduces the impact of irrelevant parts in $\mathbf M$, resulting in a more focused attention through $\hat{\mathbf M}$. }
\label{fig:qualitative_results}
\end{figure*}


Figure \ref{fig:ablation_temporal_spatial}(a) and (b) illustrate the impact of $r_{min}$ and $r_{max}$ on the performance, respectively. In the in-dataset scenario, no significant difference is observed when adjusting $r_{min}$ and $r_{max}$. However, in the cross-dataset scenario, high values of $r_{min}$ notably degrade the performance, emphasizing the importance of temporal locality achieved through the generation of pseudo-fake data. Conversely, low values of $r_{max}$ lead to a decreased performance, highlighting the importance of diversity in the pseudo-fake data.

\subsection{Comparisons with state-of-the-art}

Table \ref{tab:cross_soa} and Table \ref{tab:in_soa}  compare the proposed approach with SoA in terms of AUC on cross-dataset and in-dataset settings, respectively. For this comparison, we use the best performing model with $r_{min} = 0.3$ and $r_{max} = 1$. It can be seen that in general methods that are based on two modalities outperform single-modality methods. We also note that our method achieves better performance than audio-visual techniques, including, inconsistency-based methods that use more global features in the visual data, e.g., MDS \cite{chugh2020not} and Emotion \cite{mittal2020emotions}.

\subsection{Qualitative results}
For a deeper understanding of the proposed method, we show in Figure \ref{fig:qualitative_results} the distance map $\mathbf M$, attention map $\mathbf A$, and the attention-aware distance map $\hat{\mathbf M}$  extracted from several samples of the FakeAVCeleb dataset. As observed, $\mathbf M$ may occasionally exhibit high distances in irrelevant parts, such as the background. However, $\mathbf A$ reduces the impact of some irrelevant zones, allowing the consideration of more important portions. This trend is observed in both fake and real data. It can also be noted that the highlighted distance can look similar for both real and fake data. However, as observed in Figure \ref{fig:ablation_meandistance_distribution}(b), the overall distance map is different, creating a separation between real and fake data.


\section{Conclusion}
\label{sec:conclusion}


 In this paper, a novel fined-grained method for audio-visual deep detection is proposed. Instead of measuring the inconsistency between global audio and visual features, more local strategies are considered. First, we propose a spatially-local architecture that computes the inconsistency between relevant visual patches  and audio. Second, a temporally-local pseudo-fake data synthesis is introduced. The generated pseudo-fakes incorporating subtle inconsistencies are then used for training the proposed architecture. Experiments demonstrate the importance of the proposed components and their competitiveness as compared to the SoA. Several future directions include exploring different design choices and datasets.

 \section*{Acknowledgment}
This work was supported by the Luxembourg National Research Fund (FNR) under the project BRIDGES2021/IS/16353350/FaKeDeTeR and POST Luxembourg.
\bibliography{egbib}

\appendix

\include{bmvc_arxiv_supplementary}

\end{document}

%% file: bmvc_arxiv_supplementary.tex
\maketitle

\begin{abstract}
This supplementary material accompanies the paper titled "Detecting Audio-Visual Deepfakes with Fine-Grained Inconsistencies" and includes additional illustrations of our method along with further qualitative results.
\end{abstract}

\section{More illustrations on the method}

\begin{figure*}[]
\centering
\includegraphics[width=0.5\linewidth]{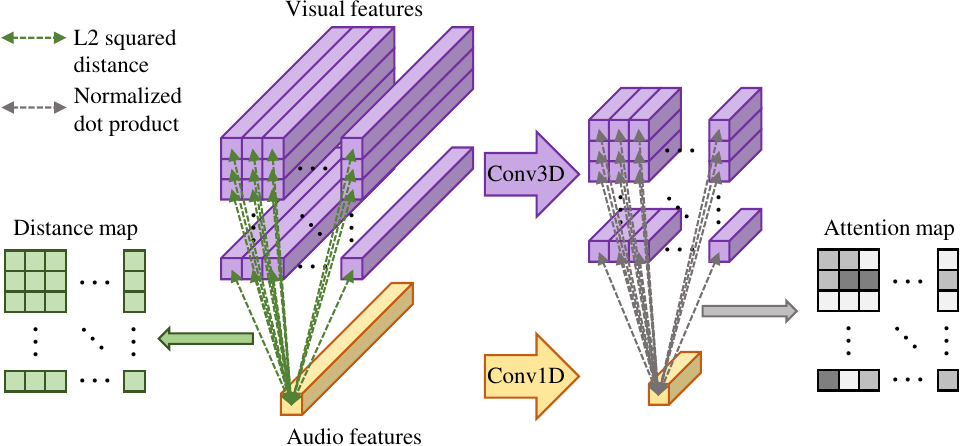}
\vspace{3mm}
\caption{Calculation of the distance map (Section 3.1.2 of the manuscript) and attention map (Section 3.1.3 of the manuscript) based on the visual and audio features.}
\label{fig:detail_cross_att_and_distance}
\end{figure*}

\begin{figure*}[]
\centering
\includegraphics[width=0.5\linewidth]{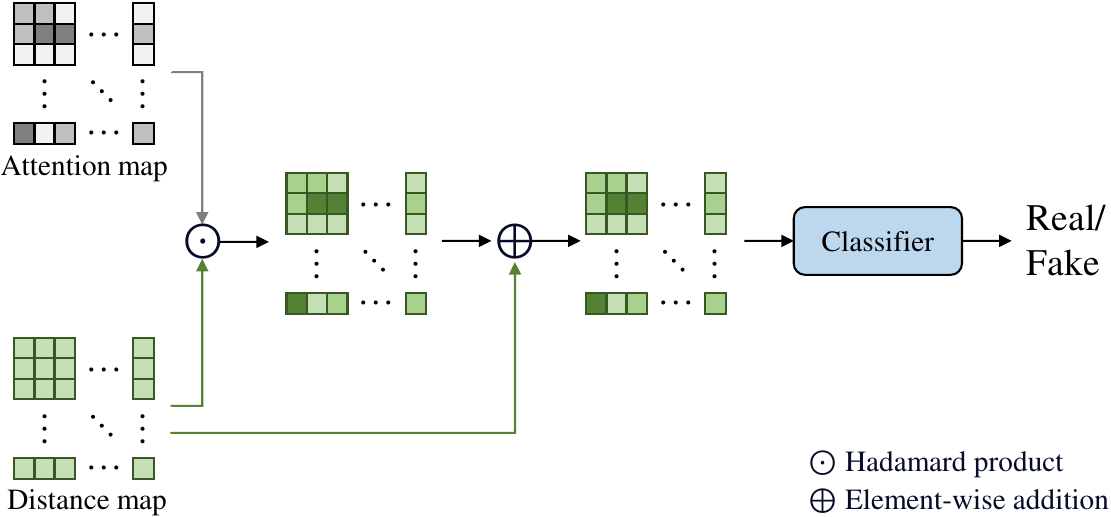}
\vspace{3mm}
\caption{Model with residual connection used in the ablation study.(Section 4.2.1 of the manuscript).}
\label{fig:method_residual_conn}
\end{figure*}

In this section, we provide additional figures to assist readers in understanding the method outlined in the main manuscript. Figure \ref{fig:detail_cross_att_and_distance} demonstrates the calculation of the distance map and attention map as described in Section 3.1.2 (Spatially-local distance map) and 3.1.3 (Attention module), respectively. Additionally, Figure \ref{fig:method_residual_conn} illustrates the model with residual connection utilized in the ablation study discussed in Section 4.2.1 of the manuscript.

\section{More qualitative results}

\begin{figure}[tb]
\centering
\begin{minipage}{.47\textwidth}
  \centering
  \includegraphics[width=\linewidth]{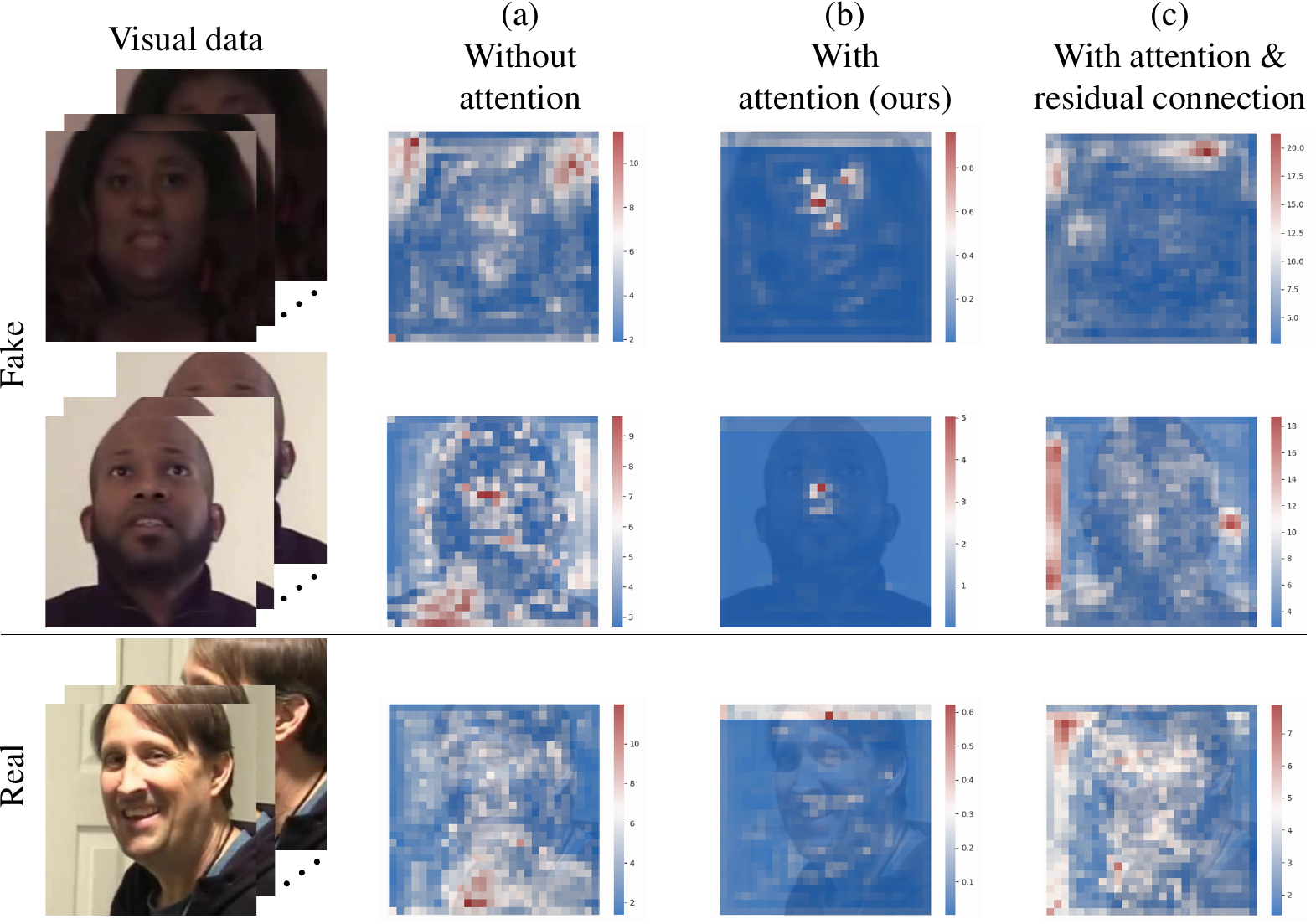}
  \vspace{2mm}
  \captionof{figure}{Visualization of $\hat{\mathbf M}$ on a few samples from the DFDC test split. This figure corresponds to Figure 6 of the main manuscript.}
  \label{fig:dfdc_ablation_qualitative}
\end{minipage}%
\hfill
\begin{minipage}{.50\textwidth}
  \centering
  \includegraphics[width=0.75\linewidth]{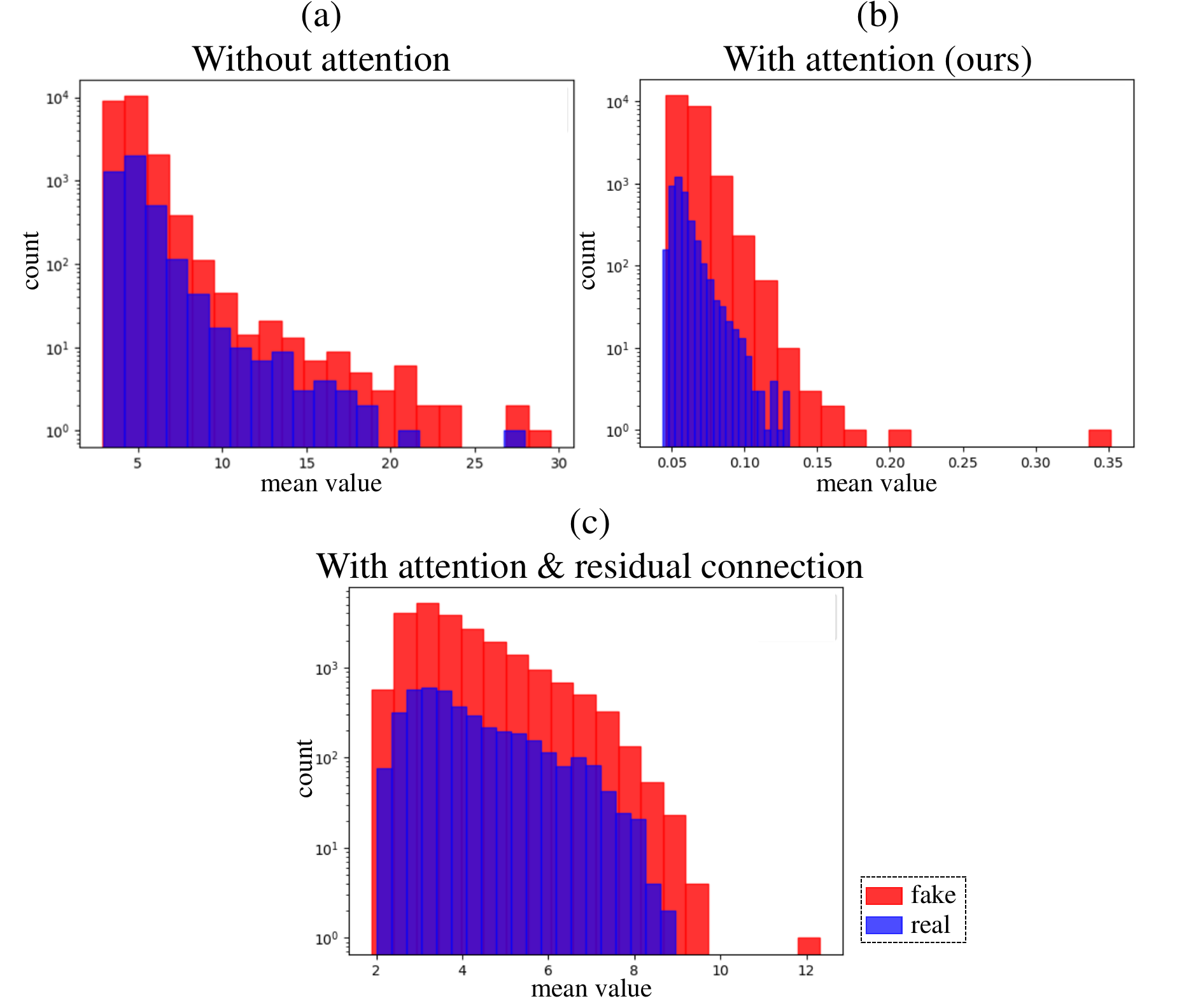}
  \vspace{2mm}
  \captionof{figure}{Histograms illustrating the distribution of the mean value of $\hat{\mathbf M}$ on the real and fake data of the DFDC test split. This figure corresponds to Figure 7 of the main manuscript.}
  \label{fig:dfdc_ablation_meandistance_distribution}
\end{minipage}
\end{figure}
\begin{figure*}[h]
\centering
\includegraphics[width=\linewidth]{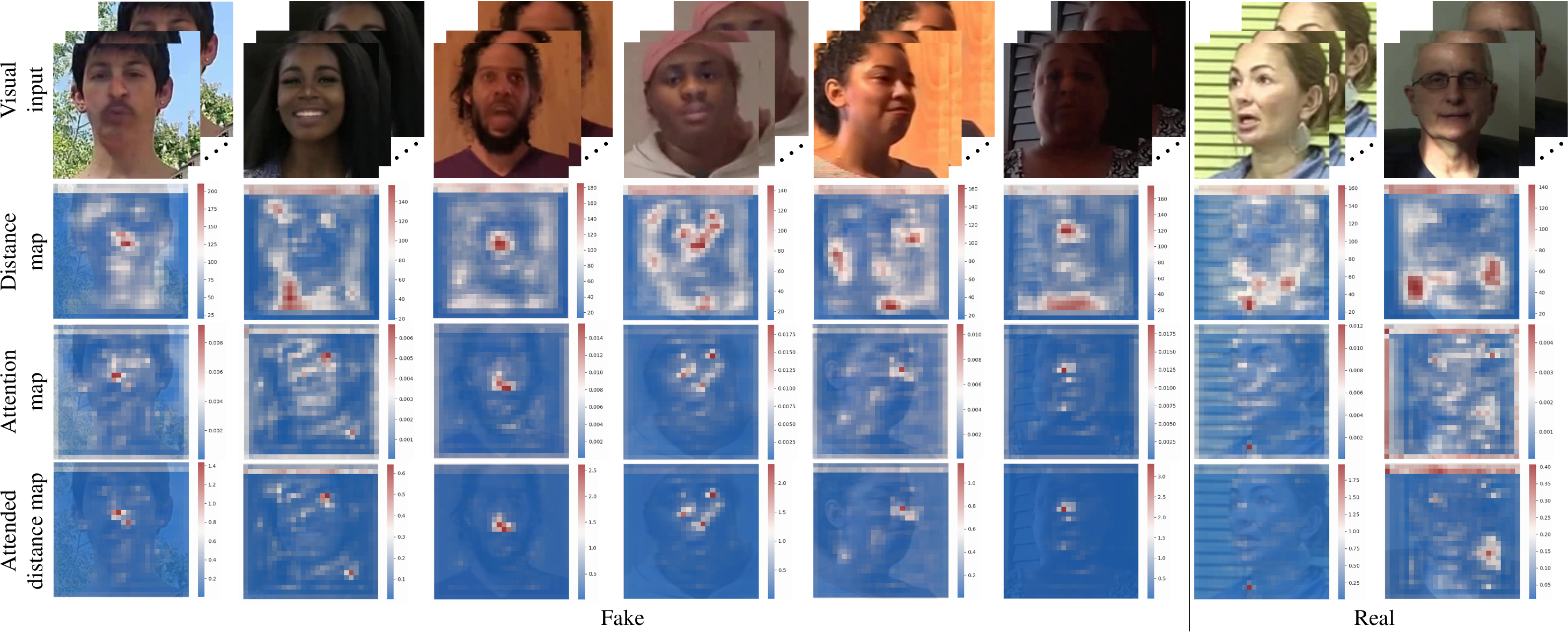}
\vspace{0.5mm}
\caption{Visualization of the distance map $\mathbf M$, attention map $\mathbf A$, and the attended distance map $\hat{\mathbf M}$ for several examples from the DFDC dataset. This figure corresponds to Figure 8 of the main manuscript.}
\vspace{-1mm}
\label{fig:dfdc_qualitative_results}
\end{figure*}

We also present additional results on the DFDC dataset to complement the FakeAVCeleb results presented in the main manuscript. Figures \ref{fig:dfdc_ablation_qualitative}, \ref{fig:dfdc_ablation_meandistance_distribution}, and \ref{fig:dfdc_qualitative_results} correspond to Figures 6, 7, and 8 of the main manuscript, respectively. Similar observations to those with the FakeAVCeleb dataset in the main manuscript are noted. However, since the performance difference is not as significant compared to FakeAVCeleb (see Table 1(b), (d), (e) of the manuscript), the distribution difference between settings shown in Figure \ref{fig:dfdc_ablation_meandistance_distribution} is not as pronounced as the one shown in Figure 7 of the manuscript.

\begin{figure*}[h]
\centering
\includegraphics[width=0.5\linewidth]{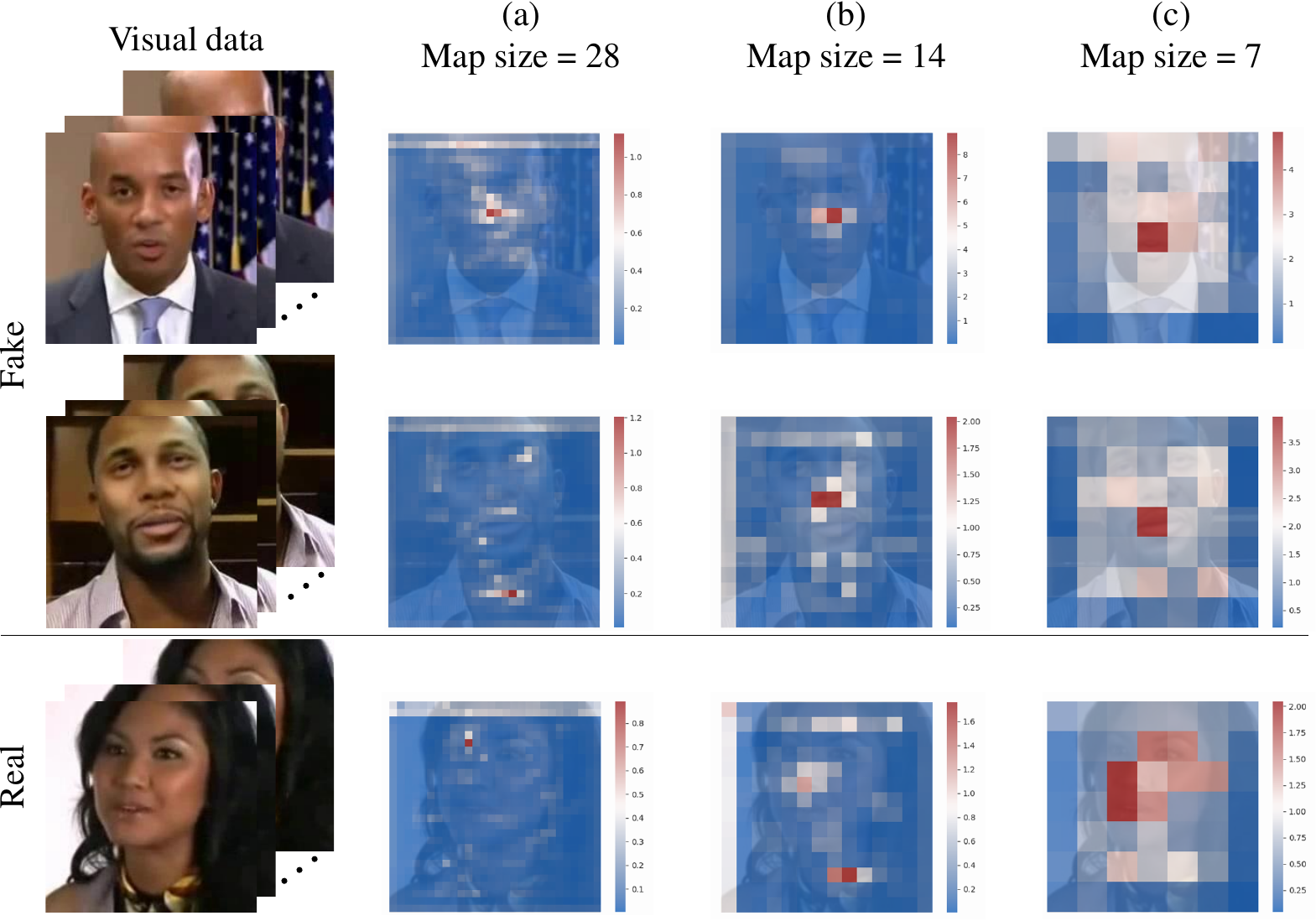}
\vspace{1mm}
\caption{Visualization of the attended distance map $\hat{\mathbf M}$ for several examples from the FakeAVCeleb dataset with different map sizes ($H^{v\prime}$ and $W^{v\prime}$).}
\vspace{-1mm}
\label{fig:ablation_spasial_qualitative}
\end{figure*}

Figure \ref{fig:ablation_spasial_qualitative} presents visualizations of $\hat{\mathbf M}$ with different map sizes, corresponding to the results reported in Figure 5(c) of the main manuscript. Despite the less fine-grained setup, our model is capable of identifying inconsistency-prone regions, resulting in a minimal performance drop (as observed in Figure 5(c) of the manuscript).

